\documentclass{article}
\usepackage{ijcai19}

\usepackage{times}
\usepackage{soul}
\usepackage{url}
\usepackage[hidelinks]{hyperref}
\usepackage[utf8]{inputenc}
\usepackage[small]{caption}
\usepackage{graphicx}
\usepackage{amsmath}
\usepackage{amssymb}
\usepackage{booktabs}
\usepackage{microtype}

\usepackage[ruled,vlined,linesnumbered]{algorithm2e}
\usepackage{mathtools}
\usepackage{multirow}

\title{Multi-view Knowledge Graph Embedding for Entity Alignment}

\author{
	Qingheng Zhang$^{1,}$\thanks{Equal contributors}\and
	Zequn Sun$^{1,*}$\and
	Wei Hu$^{1,}$\thanks{Corresponding author}\and
	Muhao Chen$^2$\and
	Lingbing Guo$^1$\and
	Yuzhong Qu$^1$\\
\affiliations
$^1$ State Key Laboratory for Novel Software Technology, Nanjing University, China\\
$^2$ Department of Computer Science, University of California, Los Angeles, USA\\
\emails
\{qhzhang, zqsun, lbguo\}.nju@gmail.com,
\{whu, yzqu\}@nju.edu.cn, 
muhaochen@ucla.edu
}

\begin{document}

\maketitle


\begin{abstract}
We study the problem of embedding-based entity alignment between knowledge graphs (KGs). Previous works mainly focus on the relational structure of entities. Some further incorporate another type of features, such as attributes, for refinement. However, a vast of entity features are still unexplored or not equally treated together, which impairs the accuracy and robustness of embedding-based entity alignment. In this paper, we propose a novel framework that unifies multiple views of entities to learn embeddings for entity alignment. Specifically, we embed entities based on the views of entity names, relations and attributes, with several combination strategies. Furthermore, we design some cross-KG inference methods to enhance the alignment between two KGs. Our experiments on real-world datasets show that the proposed framework significantly outperforms the state-of-the-art embedding-based entity alignment methods. The selected views, cross-KG inference and combination strategies all contribute to the performance improvement.
\end{abstract}


\section{Introduction}
\label{sect:intro}

\emph{Entity alignment}, a.k.a.~entity matching or resolution, aims to find entities in different knowledge graphs (KGs) referring to the same real-world identity. It plays a fundamental role in KG construction and fusion, and also supports many downstream applications, e.g., semantic search, question answering and recommender systems. Conventional methods for entity alignment identify similar entities based on the symbolic features, such as names, textual descriptions and attribute values. However, the computation of feature similarity often suffers from the semantic heterogeneity between different KGs~\cite{ActiveGenLink}. Recently, increasing attention has been paid to leveraging the KG embedding techniques for addressing this problem, where the key idea is to learn vector representations (called \emph{embeddings}) of KGs and find entity alignment according to the similarity of the embeddings. The vector representations can benefit the task of learning similarities~\cite{StarSpace}. Although existing embedding-based entity alignment methods have achieved promising results, they are still challenged by the following two limitations.

First, entities in KGs have various features, but the current embedding-based entity alignment methods exploit just one or two types of them. For example, MTransE \cite{MTransE}, IPTransE \cite{IPTransE} and BootEA \cite{BootEA} only embed the relational structure of KGs for entity alignment, in addition to which JAPE~\cite{JAPE}, KDCoE~\cite{KDCoE} and AttrE \cite{AttrE} complement attributes, textual descriptions or literals, respectively, to refine the embeddings. Actually, different types of features characterize different aspects of entity identities. Making use of them together would improve the accuracy and robustness.

Second, the existing embedding-based entity alignment methods rely on abundant \emph{seed} entity alignment as labeled training data. However, in practice, such seed entity alignment is not always accessible and very costly to obtain \cite{ActiveGenLink}. Furthermore, it is widely-acknowledged that entity alignment can benefit from relation and attribute alignment \cite{PARIS}. Still, the existing methods \cite{IPTransE,AttrE} assume that seed relation and attribute alignment can be easily found beforehand. In fact, learning embeddings from various features would enable the automatic population of more alignment information and relieve the reliance on the seed alignment.

\begin{figure}
	\centering
	\includegraphics[width=\columnwidth]{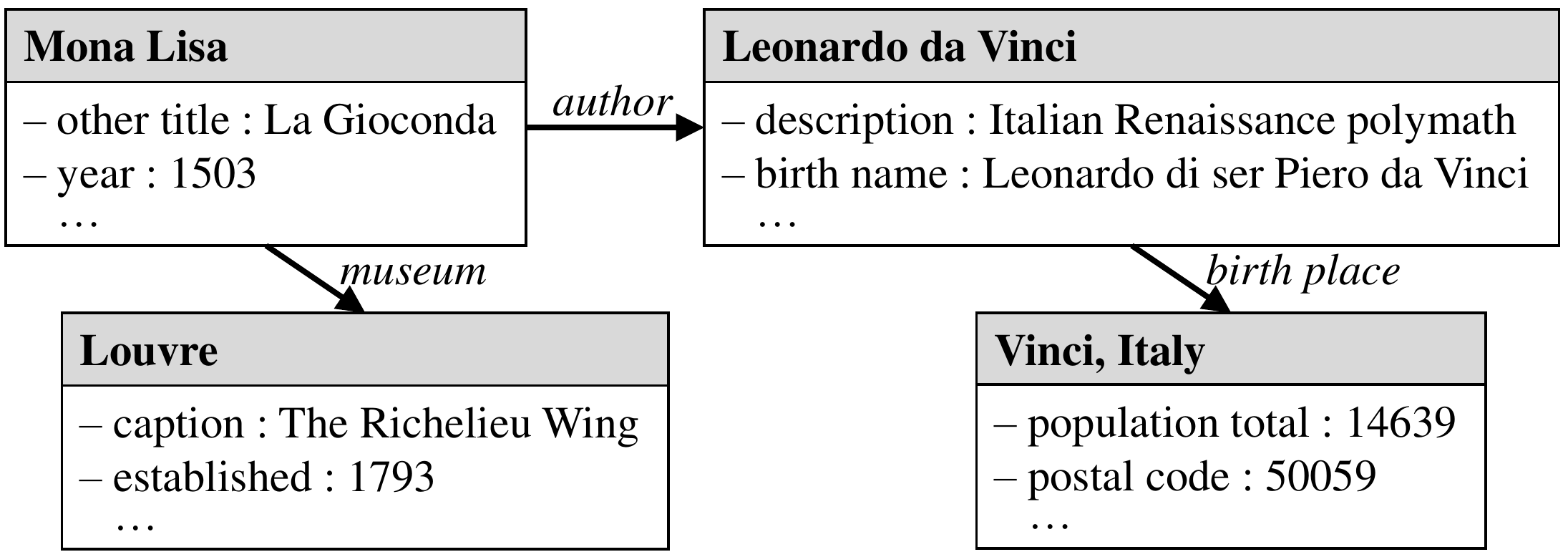}
	\caption{An example of the multi-view features, e.g., names (denoted by \textbf{bold} font), relations (denoted by \textit{italic} font) and attributes (denoted by regular font), of four entities in DBpedia.}
	\label{fig:example}
\end{figure}

To cope with the above limitations, we propose MultiKE, a new entity alignment framework based on multi-view KG embedding. The underlying idea is to divide the various features of KGs into multiple subsets (called \emph{views}), which are complementary to each other (see Figure~\ref{fig:example} for example). So entity embeddings can be learned from each particular view and jointly optimized to improve the alignment performance. In summary, our main contributions are listed as follows:
\begin{itemize}
\item Based on the data model of KGs, we define three representative views based on the name, relation and attribute features. For each view, we employ an appropriate model to learn embeddings from it. (Section~\ref{sect:multiview})

\item For entity alignment, we design two cross-KG identity inference methods at the entity level as well as the relation and attribute level, respectively, to preserve and enhance the alignment between different KGs. (Section~\ref{sect:cross})

\item We present three different strategies to combine multiple view-specific entity embeddings. Finally, we find entity alignment by the combined embeddings. (Section~\ref{sect:comb})

\item Our experiments on two real-world datasets show that MultiKE largely outperforms existing embedding-based entity alignment methods. The selected views, cross-KG inference and combination strategies all contribute to the improvement. MultiKE also achieves promising results on unsupervised entity alignment and is comparable to conventional entity alignment methods.  (Section~\ref{sect:exp})
\end{itemize}


\section{Related Work}
\label{sect:work}


\paragraph{KG embedding.} Learning KG embeddings has drawn much attention in recent years. Current KG embedding models can be divided into three kinds. \emph{Translational} models, such as TransE \cite{TransE}, TransH \cite{TransH} TransR \cite{TransR} and TransD~\cite{TransD}, interpret a relation as a translation vector from a head entity to a tail. \emph{Semantic matching} models use similarity-based functions to infer relation facts, e.g., the Hadamard product in DistMult~\cite{DistMult} and ComplEx \cite{ComplEx}, and the circular correlation in HolE \cite{HolE}. \emph{Neural} models exploit deep learning techniques for KG embedding, e.g., multilayer perceptrons in ProjE \cite{ProjE}, convolutional neural networks in ConvE \cite{ConvE}, and graph convolutional networks in R-GCN~\cite{RGCN}. All these models focus on relation facts and are mostly evaluated by the task of link prediction in a single KG. 

\paragraph{Embedding-based entity alignment.} The current methods represent different KGs as embeddings and find entity alignment by measuring the similarity between these embeddings. MTransE \cite{MTransE} learns a mapping between two separate KG embedding spaces. IPTransE \cite{IPTransE} and BootEA \cite{BootEA} are two self-training methods, which embed two KGs in a unified space and iteratively label new entity alignment as supervision. GCN-Align \cite{GCNAlign} employs graph convolutional networks to model entities based on their neighborhood information. These methods align entities mostly based on the relation features of KGs. Additionally, several other methods leverage an additional type of features to enhance the relation-based embeddings. JAPE \cite{JAPE} models the attribute correlations with Skip-Gram, KDCoE \cite{KDCoE} co-trains the embeddings of textual descriptions, and AttrE \cite{AttrE} conducts character-level literal embeddings. However, all these features are not equally treated but used to refine the relation-based embeddings. More importantly, these methods are incapable of incorporating new features.

\paragraph{Multi-view representation learning.} Learning representations from multi-view data can achieve strong generalization performance. Recently, multi-view representation learning has been widely used in the network embedding~\cite{MVE,I2MNE} and natural language processing (NLP)~\cite{CVT}. A typical process of multi-view representation learning is constituted by three major steps: (i) identify multiple views that can sufficiently represent the data; (ii) carry out representation learning on each view; and (iii) combine the multiple view-specific representations. We refer interested readers to \cite{MVSurvey} for more details. 


\section{Multi-view KG Embedding}
\label{sect:multiview}

\subsection{Problem Statement}
In this paper, we study multi-view KG embedding for entity alignment, which aims to learn comprehensive entity embeddings based on different views. Particularly, for entity alignment, we consider three views for an entity identity, namely the name view, the relation view and the attribute view. Names describe a basic and key characteristic of entities. In KGs, entity names are usually described, for example, by \textit{rdfs:label}, or the local names of entity URIs\footnote{A local name is the string after the last hash or slash of a URI.}. The relations and attributes defined in KG schemata characterize entities with rich extrinsic and intrinsic information. 
Based on the view division above, we formalize a KG as a 7-tuple $\mathcal{G}=(\mathcal{E},\mathcal{R},\mathcal{A},\mathcal{V},\mathcal{N},\mathcal{X}
,\mathcal{Y})$, where $\mathcal{E},\mathcal{R},\mathcal{A}$ and $\mathcal{V}$ denote the sets of entities, relations, attributes and literals, respectively. $\mathcal{N}\subseteq\mathcal{E}\times\mathcal{V}$ denotes the name view of entities, $\mathcal{X}\subseteq\mathcal{E}\times\mathcal{R}\times\mathcal{E}$ denotes the relation view, and $\mathcal{Y}\subseteq\mathcal{E}\times\mathcal{A}\times \mathcal{V}$ denotes the attribute view. 

Given a source KG $\mathcal{G}_a=(\mathcal{E}_a,\mathcal{R}_a,\mathcal{A}_a,\mathcal{V}_a,\mathcal{N}_a,\mathcal{X}_a,\mathcal{Y}_a)$ and a target KG $\mathcal{G}_b=(\mathcal{E}_b,\mathcal{R}_b,\mathcal{A}_b,\mathcal{V}_b,\mathcal{N}_b,\mathcal{X}_b,\mathcal{Y}_b)$, entity alignment aims to find a set of identical entities $\mathcal{M}=\{(e_i,e_j)\in \mathcal{E}_a\times\mathcal{E}_b \mid e_i\equiv e_j\}$, where ``$\equiv$'' denotes the equivalence relationship. We first learn $d$-dimensional entity embeddings and then find entity alignment based on the embeddings.

For notations, we use bold lowercase letters to represent embeddings and bold uppercase letters to matrices. The superscript with bracket of an entity embedding indicates from which view this embedding comes. The name, relation and attribute views are marked by $^{(1)},^{(2)}$ and $^{(3)}$, respectively.


\subsection{Literal Embedding}
Literals are constituted by sequences of tokens. As the basis for multi-view embedding, literal embedding represents the discrete and symbolic literals by $d$-dimensional embeddings.  Without loss of generality, let $l = (o_1, o_2, \ldots, o_n) $ denote a literal of $n$ tokens. $\text{LP}(\cdot)$ is defined as a lookup function that maps the input to an embedding. To resolve the different expressions of literals and capture their inherent semantics, we consider the following cases:
\begin{align}\small
\text{LP}(o_i) = 
\begin{cases}
\text{word\_embed}(o_i) & \text{if } o_i \text{ has a word embedding} \\
\text{char\_embed}(o_i) & \text{otherwise}
\end{cases},
\end{align}
where $\text{word\_embed}(\cdot)$ returns the corresponding word embedding of the input token. In fact, word embeddings have become a fundamental resource for many NLP applications, due to their capability of capturing semantic relatedness of words. Here, we use pre-trained word embeddings~\cite{FastText} to collocate similar literals. $\text{char\_embed}(\cdot)$ returns the average of the character embeddings that are pre-trained with the Skip-Gram model \cite{SkipGram} on the KG literal set $\mathcal{V}_a \cup \mathcal{V}_b$. Let $\phi(\cdot)$ be the literal embedding of the input. We employ an autoencoder to encode a list of token embeddings into one literal embedding in an unsupervised manner:
\begin{align}
\phi(l) = \text{encode}([\text{LP}(o_1); \text{LP}(o_2); \ldots; \textsc{LP}(o_n)]),
\end{align}
where $\text{encode}(\cdot)$ returns the compressed representation of the input embeddings and $[;]$ denotes the concatenation operation. We limit the maximum number of tokens to 5. Long literals are truncated while short ones are appended with placeholders.


\subsection{Name View Embedding}
We embed the name view using the above literal embeddings. Given an entity $h$, its name embedding is defined as follows:
\begin{align}
\mathbf{h}^{(1)} = \phi(\text{name}(h)),
\end{align}
where $\text{name}(\cdot)$ extracts the name of the input object. We refer to the KG embeddings under the name view as $\Theta^{(1)}$. 


\subsection{Relation View Embedding}
The relation view characterizes the structure of KGs, in which entities are linked by relations. To preserve such relational structures, we adopt TransE~\cite{TransE} to interpret a relation as a translation vector from its head entity to tail entity. Given a relation fact $(h,r,t)$ in KGs, we use the following score function to measure the plausibility of the embeddings:
\begin{align}
f_{\text{rel}}(\mathbf{h}^{(2)}, \mathbf{r}, \mathbf{t}^{(2)}) = - ||\mathbf{h}^{(2)} + \mathbf{r} - \mathbf{t}^{(2)}||,
\end{align}
where $||\cdot||$ denotes either $L_1$ or $L_2$ vector norm. Then, we define the probability of $(h, r, t)$ being a real relation fact (i.e., existing in the KGs) as follows:
\begin{align}
\label{eq:rel}
\resizebox{.91\columnwidth}{!}{$\displaystyle
\mathcal{P}_{\text{rel}}(\zeta_{(h,r,t)}=1\,|\,\Theta^{(2)}) = \text{sigmoid}(f_{\text{rel}}(\mathbf{h}^{(2)}, \mathbf{r}, \mathbf{t}^{(2)})),
$}
\end{align}
where $\Theta^{(2)}$ denotes the KG embeddings under the relation view and  $\zeta_{(h,r,t)}$ denotes the label (``1" or ``$-1$") of $(h,r,t)$. We parameterize $\Theta^{(2)}$ by minimizing the logistic loss below:
\begin{align}
\resizebox{.91\columnwidth}{!}{$\displaystyle
\mathcal{L}(\Theta^{(2)}) = \sum_{\mathclap{(h,r,t)\in\mathcal{X}^{+}\cup\mathcal{X}^{-}}} \log( 1+\exp( -\zeta_{(h,r,t)}f_{\text{rel}}(\mathbf{h}^{(2)},\mathbf{r},\mathbf{t}^{(2)}))),
$}
\end{align}
where $\mathcal{X}^{+}=\mathcal{X}_a \cup \mathcal{X}_b$ denotes the set of real relation facts in the source and target KGs, while $\mathcal{X}^{-}$ denotes the set of faked ones sampled by replacing the head or tail entities of real relation facts with random ones.


\subsection{Attribute View Embedding}
For the attribute view, we use a convolutional neural network (CNN) to extract features from the attributes and values of entities. We splice the embeddings of an attribute $a$ and its value $v$ into a matrix, denoted by $\langle \mathbf{a}; \mathbf{v}\rangle \in \mathbb{R}^{2\times d}$, and feed it to a CNN to obtain the compressed representation:
\begin{align}
\text{CNN}(\langle\mathbf{a}; \mathbf{v}\rangle) = \sigma(\text{vec}(\sigma(\langle\mathbf{a}; \mathbf{v}\rangle*\Omega))\mathbf{W}),
\end{align}
where $\text{CNN}(\cdot)$ denotes a convolution operation that slides a convolution kernel $\Omega$ of size $2\times c$ ($c < d$) over the input embedding matrix to extract high-level features. The feature map tensor is then reshaped to a vector by $\text{vec}(\cdot)$ and finally projected in the KG embedding space using a densely-connected layer parametrized by $\mathbf{W}$. $\sigma(\cdot)$ is an activation function. 

Given an attribute fact $(h,a,v)$ in KGs, we define the following score function to measure its plausibility:
\begin{align}
f_{\text{attr}}(\mathbf{h}^{(3)}, \mathbf{a}, \mathbf{v}) = -||\mathbf{h}^{(3)} - \text{CNN}(\langle\mathbf{a}; \mathbf{v}\rangle)||.
\end{align}

Based on this, the head entities are expected to be close to its attributes and values. This objective can be achieved by minimizing the following logistic loss: 
\begin{align}
	\mathcal{L}(\Theta^{(3)}) = \sum_{\mathclap{(h,a,v)\in\mathcal{Y}^{+}}} \log( 1+\exp( -f_{\text{attr}}(\mathbf{h}^{(3)}, \mathbf{a}, \mathbf{v}))),
\end{align}
where $\mathcal{Y}^{+}=\mathcal{Y}_a \cup \mathcal{Y}_b$ denotes the set of real attribute facts in the source and target KGs, and $\Theta^{(3)}$ denotes the KG embeddings under the attribute view. Negative sampling is not employed here, because we find that it does not lead to noticeable improvement on entity alignment.

\section{Cross-KG Training for Entity Alignment}
\label{sect:cross}

In this section, we propose the cross-KG entity identity inference to capture the alignment information between two KGs, based on seed entity alignment. Furthermore, we present the cross-KG relation and attribute identity inference to enhance entity alignment, based on the automatically found relation and attribute alignment.

\begin{figure}
	\centering
	\includegraphics[width=\columnwidth]{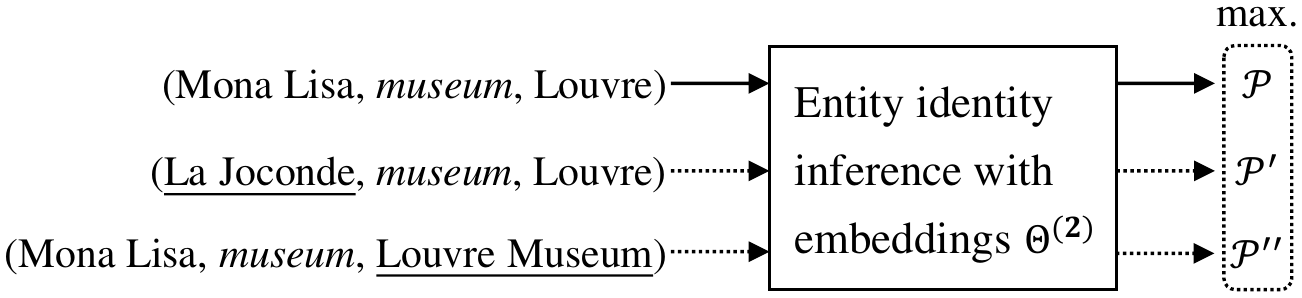}
	\caption{Illustration of cross-KG entity identity inference under the relation view, where entities without and with \underline{underlines} come from DBpedia and Wikidata, respectively. Dotted arrows denote auxiliary inference probabilities.}
	\label{fig:cross_kg}
\end{figure}

\subsection{Entity Identity Inference}
As shown in Figure~\ref{fig:cross_kg}, we take as an example the cross-KG entity identity inference under the relation view. Our intuition is that swapping aligned entities in their involved relation facts should lead to the same probability of identity inference, because the entities refer to the same object in the real world~\cite{IPTransE,BootEA}. Given a relation fact $(h,r,t)$, if $(h,\hat{h})$ appears in the seed entity alignment, we add the following auxiliary probability:
\begin{align}
\resizebox{.89\columnwidth}{!}{$\displaystyle
\mathcal{P}'_{\text{rel}}(\zeta_{(h,r,t)}=1\,|\,\Theta^{(2)}) = \text{sigmoid}(f_{\text{rel}}(\hat{\mathbf{h}}^{(2)}, \mathbf{r}, \mathbf{t}^{(2)})).
$}
\end{align}

Also, if $(t,\hat{t})$ exists in the seed entity alignment, we add:
\begin{align}
\resizebox{.89\columnwidth}{!}{$\displaystyle
\mathcal{P}''_{\text{rel}}(\zeta_{(h,r,t)}=1\,|\,\Theta^{(2)}) = \text{sigmoid}(f_{\text{rel}}(\mathbf{h}^{(2)},\mathbf{r},\hat{\mathbf{t}}^{(2)})).
$}
\end{align}

We maximize these auxiliary probabilities over the relation facts having those entities in the seed entity alignment. The loss is computed as follows:
\begin{align}
\label{eq:loss_ce}\small
\mathcal{L}_{\text{CE}}(\Theta^{(2)}) & = \sum_{\mathclap{(h,r,t)\in\mathcal{X}'}} \log(1+\exp(-f_{\text{rel}}(\hat{\mathbf{h}}^{(2)},\mathbf{r},\mathbf{t}^{(2)}))) \nonumber\\
+ & \sum_{\mathclap{(h,r,t)\in\mathcal{X}''}} \log(1+\exp(-f_{\text{rel}}(\mathbf{h}^{(2)},\mathbf{r},\hat{\mathbf{t}}^{(2)}))),
\end{align}
where $\mathcal{X}'$ and $\mathcal{X}''$ refer to the sets of relation facts whose head and tail entities are in the seed entity alignment, respectively.

For the cross-KG entity inference under the attribute view, the loss, denoted by $\mathcal{L}_{\text{CE}}(\Theta^{(3)})$, is similarly defined:
\begin{align}
\resizebox{.89\columnwidth}{!}{$\displaystyle
\mathcal{L}_{\text{CE}}(\Theta^{(3)}) = \sum_{\mathclap{(h,a,v)\in\mathcal{Y}'}} \log(1+\exp(-f_{\text{attr}}(\hat{\mathbf{h}}^{(3)},\mathbf{a},\mathbf{v}))), 
$}
\end{align}
where $\mathcal{Y}'$ denotes the set of attribute facts such that the head entities are in the seed entity alignment.


\subsection{Relation and Attribute Identity Inference}
Similar to the cross-KG entity identity inference, we add the auxiliary probabilities for the cross-KG relation and attribute identity inference. Let us consider the relations for example. Given a relation fact $(h,r,t)$, if $(r,\hat{r})$ constitutes a relation alignment, we have:
\begin{align}
\resizebox{.89\columnwidth}{!}{$\displaystyle
\mathcal{P}'''_{\text{rel}}(\zeta_{(h,r,t)}=1\,|\,\Theta^{(2)}) = \text{sigmoid}(f_\text{rel}(\mathbf{h}^{(2)},\hat{\mathbf{r}},\mathbf{t}^{(2)})).
$}
\end{align}

Unlike the previous works \cite{IPTransE,AttrE} that assume the existence of seed relation and attribute alignment, we propose a \emph{soft} alignment method to automatically find the relation and attribute alignment in training. Here, ``soft'' means that we do not require the relations or attributes in the alignment to be strictly equivalent. In fact, due to the heterogeneity between KG schemata, finding relation and attribute alignment can be harder than entity alignment \cite{PropAlign}, and sometimes the strict alignment even does not exist. We denote the soft relation alignment by $\mathcal{S}_{\text{rel}}=\{(r,\hat{r},\text{sim}(r,\hat{r}))\,|$ $\text{sim}(r,\hat{r})\geq\eta\}$, where $r,\hat{r}$ are relations from different KGs, $\text{sim}(r,\hat{r})$ denotes their similarity, and $\eta$ denotes a threshold in $(0,1]$. We consider two kinds of similarity: name similarity based on literal embeddings and semantic similarity based on relation embeddings. We combine them as a weighted sum:
\begin{align}
\resizebox{.89\columnwidth}{!}{$\displaystyle
\text{sim}(r,\hat{r})=\alpha_1\cos(\phi(\text{name}(r)),\phi(\text{name}(\hat{r})))+\alpha_2\cos(\mathbf{r},\hat{\mathbf{r}}),
$}
\end{align}
where $\alpha_1,\alpha_2>0$ are two weighting factors and $\alpha_1+\alpha_2=1$. $\cos(\cdot)$ calculates the cosine similarity of two embeddings. 

We regard this similarity as a smooth coefficient to reduce the negative effect of inaccurate alignment and combine it into the loss of cross-KG relation identity inference:
\begin{align}
\resizebox{.89\columnwidth}{!}{$\displaystyle
\mathcal{L}_{\text{CRA}}(\Theta^{(2)}) = \sum_{\mathclap{(h,r,t) \in \mathcal{X}'''}} \text{sim}{(r,\hat{r})} \log(1+\exp(-f_{\text{rel}}(\mathbf{h}^{(2)},\hat{\mathbf{r}},\mathbf{t}^{(2)}))),
$}
\end{align}
where $\mathcal{X}'''$ denotes the set of relation facts having the relations in $\mathcal{S}_{\text{rel}}$. Note that, the soft relation alignment is not fixed but updated iteratively during the training process of MultiKE.

The cross-KG attribute identify inference can be formulated in the same way. Due to the space limitation, we omit its loss function $\mathcal{L}_{\text{CRA}}(\Theta^{(3)})$ here.


\section{View Combination}
\label{sect:comb}

The view-specific embeddings characterize entity identities from different aspects. Intuitively, the general entity embeddings can benefit from  multiple view-specific embeddings. In this section, we present three combination strategies.

\subsection{Weighted View Averaging}
A straightforward combination is to average the embeddings from different views. To emphasize on important views, we assign weights to view-specific entity embeddings. Let $\tilde{\mathbf{h}}$ denote the combined embedding for $h$. Without loss of generality, let $D$ be the number of views, and we have $\tilde{\mathbf{h}} = \sum_{i=1}^{D} w_i\mathbf{h}^{(i)}$, where $w_i$ is the weight of $\mathbf{h}^{(i)}$, and can be calculated by:
\begin{align}
w_i = \frac{\cos(\mathbf{h}^{(i)}, \bar{\mathbf{h}})}{\sum_{j=1}^{D}\cos(\mathbf{h}^{(j)}, \bar{\mathbf{h}})},
\end{align}
where $\bar{\mathbf{h}}$ is the average of multi-view embeddings of $h$, i.e., $\bar{\mathbf{h}} = \frac{1}{D}\sum_{j=1}^{D}\mathbf{h}^{(j)}$. If the embedding from one view is far away from its average embedding, it would have a lower weight. This is a kind of late combination, because it aggregates embeddings after they have been learned independently.

\subsection{Shared Space Learning} 
This combination strategy seeks to induce an orthogonal mapping matrix from each view-specific embedding space to a shared space, based on the assumption that multiple views can be generated from the shared latent view. Let $\tilde{\mathbf{H}}$ be the combined embedding matrix for all entities, where each row corresponds to an entity, and $\mathbf{H}^{(i)}$ be the entity embedding matrix under the $i^\text{th}$ view. We minimize the mapping loss:
\begin{align}
\resizebox{.89\columnwidth}{!}{$\displaystyle
	\mathcal{L}_\text{SSL}(\tilde{\mathbf{H}}, \mathbf{Z}) = \sum_{i=1}^{D} (||\tilde{\mathbf{H}} - \mathbf{H}^{(i)}\mathbf{Z}^{(i)}||_F^2 + ||\mathbf{I} - \mathbf{Z}^{(i)\top}\mathbf{Z}^{(i)}||_F^2), 
$}
\end{align}
where $\mathbf{Z}=\bigcup_{i=1}^{D}\mathbf{Z}^{(i)}$. $\mathbf{Z}^{(i)}$ serves as the mapping from the $i^\text{th}$ view-specific embedding space to the shared space, and $||\cdot||_F$ denotes the Frobenius norm. The second term is used as a soft constraint to orthogonalize mapping matrices, where $\mathbf{I}$ is the identity matrix. The orthogonality helps keep the distances between embeddings in the view-specific space unchanged during the transformation, and thus the shared space can learn the alignment information from the multiple embedding spaces. Note that this is also a kind of late combination.

\subsection{In-training Combination}
Unlike the above strategies, this combination participates in the joint training of multi-view embeddings, which enables the multiple views to benefit from each other. The loss is:
\begin{align}
\mathcal{L}_\text{ITC}(\tilde{\mathbf{H}}, \mathbf{H}) = \sum_{i=1}^{D} ||\tilde{\mathbf{H}} - \mathbf{H}^{(i)}||_F^2,
\end{align}
where $\mathbf{H}=\bigcup_{i=1}^{D}\mathbf{H}^{(i)}$. The goal is to maximize the agreement between the combined embeddings and the view-specific embeddings in a unified embedding space. 

\subsection{Combined Training Process} 
The complete training process is shown in Algorithm~\ref{alg:training}. The parameters are initialized with Xavier initializer and the loss functions are optimized with AdaGrad. We first train literal embeddings based on pre-train word embeddings \cite{FastText} and character embeddings. Thus, we can directly obtain entity name embeddings. Then, we train (and combine, if in-training combination is used) embeddings from other views and perform the cross-KG entity, relation and attribute identity inference alternately. The soft alignment of relations and attributes is also updated iteratively during training. Finally, we combine the multiple view-specific entity embeddings if in-training combination is not used, and find entity alignment by the nearest-neighbor search. Note that, the name embeddings are retrieved from literal embeddings and they would not be updated in the follow-up training process.

\begin{algorithm}[t]
\caption{Combined training process of MultiKE}
\label{alg:training}
{
\KwIn{$\mathcal{G}_a,\mathcal{G}_b$, word embeddings, max epochs $Q$}
	Train literal embeddings and get the name embeddings\;
	\For{$q=1,2,\ldots,Q$}{
		Minimize $\mathcal{L}(\Theta^{(2)})$ under the relation view\;
		Minimize $\mathcal{L}(\Theta^{(3)})$ under the attribute view\;
		\If{in-training combination is used}{
			Minimize $\mathcal{L}_{\text{ITC}}(\tilde{\mathbf{H}},\mathbf{H})$\;
		}
		Minimize $\mathcal{L}_{\text{CE}}(\Theta^{(2)})$ and $\mathcal{L}_{\text{CE}}(\Theta^{(3)})$\;
		Update soft alignment $\mathcal{S}_{\text{rel}}$ and $\mathcal{S}_{\text{attr}}$\;
		Minimize $\mathcal{L}_{\text{CRA}}(\Theta^{(2)})$ and $\mathcal{L}_{\text{CRA}}(\Theta^{(3)})$\;
	}
	\If{in-training combination is not used}{
		{\small Run weighted view averaging or shared space learning}\;
	}
	Find entity alignment in $\tilde{\mathbf{H}}$ by nearest-neighbor search\;
}
\end{algorithm}


\section{Experiments}
\label{sect:exp}

In this section, we report our experiments on MultiKE. The source code is accessible online\footnote{\url{https://github.com/nju-websoft/MultiKE}}.

\subsection{Datasets}
In our experiments, we reused two datasets, namely DBP-WD and DBP-YG, recently proposed in \cite{BootEA}. The two datasets were sampled from DBpedia, Wikidata and YAGO3, each of which contains 100 thousand aligned entity pairs, i.e., reference entity alignment. Each dataset provides $30\%$ reference entity alignment as seed and leaves the remaining for evaluating entity alignment performance. 


\subsection{Comparative Methods}
We compared MultiKE with seven recent embedding-based entity alignment methods, namely MTransE, IPTransE, JAPE, BootEA, KDCoE, GCN-Align and AttrE, which have been described in Section~\ref{sect:work}. We also extended TransD, HolE and ConvE for entity alignment on behalf of the three kinds of KG embedding models, because we found that their results were most competitive in our preliminary experiments. We merged the two KGs in a dataset as one by letting the two entities in each seed entity alignment have the same embedding and used these extended models to learn embeddings. We denote the MultiKE variants that employ weighted view averaging, shared space learning and in-training combination by MultiKE-WVA, MultiKE-SSL and MultiKE-ITC, respectively.


\subsection{Experimental Settings}
The following hyper-parameters were used in the experiments. Each training took $Q=200$ epochs with learning rate 0.001. For the relation view embedding, 10 negative facts were sampled for each real relation fact. For the attribute view embedding, the number of filters was 2 and the convolution kernel size was $2\times4$ (i.e., $c=4$). The activation function for the autoencoder and CNN was $\tanh(\cdot)$. For the relation and attribute identity inference, we set $\alpha_1=0.6,\alpha_2=0.4$ and $\eta=0.9$. The embedding dimension $d$ was set to $75$ for all the comparative methods. We chose Hits@$k$ ($k=1,10$), mean rank (MR) and mean reciprocal rank (MRR) as the evaluation metrics. Higher Hits@$k$ and MRR scores as well as lower MR scores indicate better performance. Note that, Hits@1 should be more preferable, and it is equivalent to precision widely-used in conventional entity alignment.

\begin{table*}
	\centering
	\setlength{\belowcaptionskip}{-5pt}
	\resizebox{0.8\textwidth}{!}{\scriptsize
			\begin{tabular}{ll|l|llrlllrl}
				\hline
				\multicolumn{2}{l|}{\multirow{2}{*}{Features}} & \multirow{2}{*}{Methods} & \multicolumn{4}{c}{DBP-WD} & \multicolumn{4}{c}{DBP-YG} \\
				\cline{4-7} \cline{8-11} & & & Hits@1 & Hits@10 & MR& MRR & Hits@1 & Hits@10 &MR& MRR \\ 
				\hline
				\multicolumn{2}{l|}{\multirow{7}{*}{Relation only}} 
				& MTransE  & 28.12$^\dagger$ & 51.95$^\dagger$ & 656 & 0.363$^\dagger$ & 25.15$^\dagger$ & 49.29$^\dagger$ & 512 & 0.334$^\dagger$ \\
				& & IPTransE  & 34.85$^\dagger$ & 63.84$^\dagger$ & 265 & 0.447$^\dagger$ & 29.74$^\dagger$ & 55.76$^\dagger$ & 158 &0.386$^\dagger$  \\
				& & BootEA  & 74.79$^\dagger$ & 89.84$^\dagger$ &109 & 0.801$^\dagger$ & 76.10$^\dagger$ & 89.44$^\dagger$ & 34 & 0.808$^\dagger$ \\
				& & GCN-Align & 47.70 & 75.96 & 1,988 & 0.577 & 60.05 & 84.14 & 299 & 0.686\\
				\cline{3-11}
				& & TransD&36.20 & 65.08 & 152 & 0.456  & 33.49 & 59.72& 114 & 0.421 \\
				& & HolE& 22.26 & 45.22 & 811 & 0.289  & 25.04 & 48.36& 629 & 0.327\\
				& & ConvE& 40.31 & 62.78 & 1,429 & 0.483  & 50.27 & 73.56 & 837 & 0.582\\
				\hline 
				\parbox[t]{2mm}{\multirow{3}{*}{\rotatebox[origin=c]{90}{Rel.\,+}}}	
				& Attr. & JAPE & 31.84$^\dagger$ & 58.88$^\dagger$ & 266 & 0.411$^\dagger$ & 23.57$^\dagger$ & 48.41$^\dagger$ & 189 & 0.320$^\dagger$ \\  
				& Textual desc. & KDCoE & 57.19 & 69.53 & 182 & 0.618 & 42.71 & 48.30& 137& 0.446 \\
				& Literals & AttrE & 38.96 & 66.77&142 & 0.487& 23.24 &42.70 &706 & 0.300 \\
				\hline 
				\multicolumn{2}{l|}{\multirow{3}{*}{Multiple views}} 
			& MultiKE-WVA   & 90.42 & 94.59 & \textbf{22}  & 0.921 & 85.92 & 94.99 & \textbf{19} & 0.891 \\
			& & MultiKE-SSL & \textbf{91.86} & \textbf{96.26} & 39  & \textbf{0.935} &82.35&93.30&21&0.862\\
			& & MultiKE-ITC & 91.45 & 95.19 & 114 & 0.928 & \textbf{88.03} & \textbf{95.32} & 35 & \textbf{0.906} \\
				\hline
					\multicolumn{11}{l}{``$^\dagger$" indicates that the results were taken from ~\cite{BootEA}. Other results were produced using their source code.}
	\end{tabular}}
	\caption{Comparison with existing embedding-based entity alignment methods}
	\label{tab:compare}
\end{table*}

\subsection{Entity Alignment Results and Analyses}
Table~\ref{tab:compare} shows the comparison results of MultiKE and other embedding-based entity alignment methods. We found that MultiKE significantly outperforms the others on all the metrics across the two datasets. For example, on DBP-WD, MultiKE-SSL achieved a Hits@1 score of $91.86\%$ with $17.07\%$ absolute gain compared to BootEA (the second best method). This is because MultiKE explores the multi-view features of entity identities, while the others like MTransE, IPTransE and BootEA only learn from relation facts. As compared with AttrE, which leverages literals for entity alignment, MultiKE also achieved superior results (e.g., $52.9\%$ improvement of Hits@1 on DBP-WD). This is due to the fact that the pre-trained word embeddings in MultiKE can better capture the semantic similarity of literals than the character embeddings used in AttrE. Furthermore, AttrE embeds entities and literals in a unified space by TransE, which would fall short of handling the diversity of attribute values. Our three variants all achieved similar results. Note that the performance of MultiKE-SSL slightly decreased on DBP-YG. We think that this is because DBP-YG has stronger relation and attribute heterogeneity. For example, their relation numbers are $302$ vs. $31$, and the attribute numbers are $334$ vs. $23$, respectively. This hinders the combination by shared space learning. 

\paragraph{Effectiveness of views.}
\label{sect:ablation}
Table~\ref{tab:views} depicts the entity alignment results based on view-specific entity embeddings when trained independently or by in-training combination. The three views all contributed to entity alignment, especially the name view. We owe it to the proposed literal embeddings, which can capture the semantic similarity of entity names. With in-training combination, the relation and attribute views benefited from the name view and also each other, thus their results improved a lot. As name embeddings are fixed, entity alignment results in the name view are the same in different combinations. This experiment indicated that entity names and word embeddings have great potentials for capturing the entity similarity.

\begin{figure}[t!]
	\centering
	\includegraphics[width=\columnwidth]{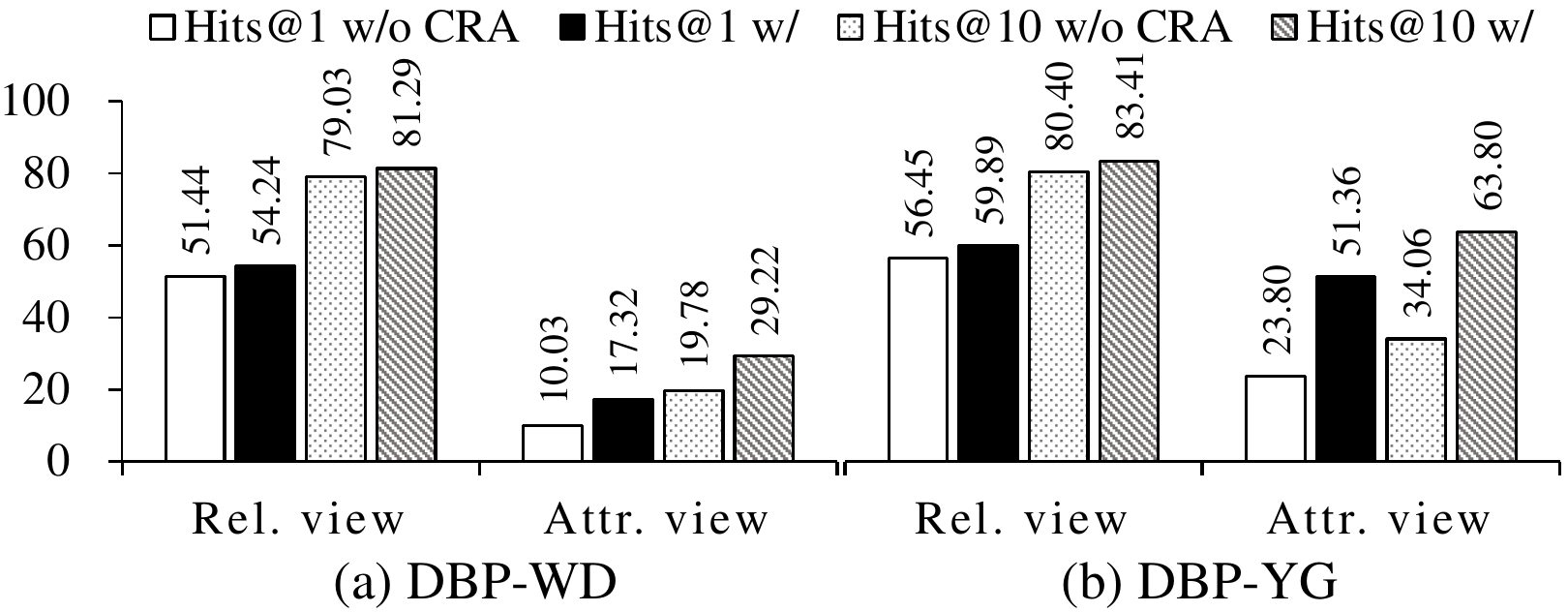}
	\caption{Results under the relation and attribute views, with or without cross-KG relation or attribute identity inference.}
	\label{fig:cross-kg}
\end{figure}

\begin{table}[t]
	\centering
	\resizebox{\columnwidth}{!}{\scriptsize
			\begin{tabular}{ll|crccrc}
				\hline & & \multicolumn{3}{c}{DBP-WD} & \multicolumn{3}{c}{DBP-YG}  \\ 
				\cline{3-5}\cline{6-8} 
				& & Hits@1 & MR& MRR & Hits@1 &MR& MRR \\
				\hline
				\parbox[t]{2mm}{\multirow{3}{*}{\rotatebox[origin=c]{90}{Indep.}}}
				& Name view  & 84.30 & 1,108 & 0.871 & 83.60 & 1,294 & 0.841 \\
				& Rel. view  & 54.24 & 169  & 0.635 & 59.89 & 58   & 0.680 \\
				& Attr. view & 17.32 & 7,688 & 0.215 & 51.36 & 1,966 & 0.558 \\ 
				\hline \parbox[t]{2mm}{\multirow{2}{*}{\rotatebox[origin=c]{90}{\scriptsize ITC}}}
				& Rel. view  & 70.22 & 34   & 0.769 & 68.57 & 47   & 0.751 \\
				& Attr. view & 61.13 & 6,774 & 0.648 & 62.52 & 1,001 & 0.688 \\ 
				\hline
	\end{tabular}}
	\caption{Results of entity alignment under independent views}
	\label{tab:views}
\end{table}
\paragraph{Effectiveness of cross-KG inference.}
Here, we examined the effectiveness of the proposed relation and attribute identity inference. We additionally performed KG embedding under the relation and attribute views without optimizing $\mathcal{L}_{\text{CRA}}(\Theta^{(2)})$ and $\mathcal{L}_{\text{CRA}}(\Theta^{(3)})$, respectively. We denote such variants by ``w/o CRA". The entity alignment results under the relation and attribute views are shown in Figure~\ref{fig:cross-kg}. We observed that the relation and attribute identity inference brought improvement to entity alignment, especially in the attribute view. This is because the attribute heterogeneity is weaker than relation heterogeneity between different KGs. The soft attribute alignment was more accurate and thus contributed more.
\paragraph{Analysis of Unsupervised Entity Alignment.}
This task aims to align entities without seed entity alignment. As found in~\cite{JAPE,GCNAlign}, seed entity alignment is vital to learn KG embeddings from relation facts. We also encountered the same issue that the relation view would fall short if no seed entity alignment is given. However, as shown in Table~\ref{tab:unsuperv}, MultiKE-ITC still achieved acceptable results, thanks to the special in-training combination. The name view does not rely on seed alignment as supervision, while the relation and attribute views can benefit from it during training. This experiment revealed that MultiKE has good robustness and can alleviate the reliance on seed alignment.
\begin{table}[!t]
\centering
\resizebox{\columnwidth}{!}{\scriptsize
	\begin{tabular}{l|crccrc}
		\hline & \multicolumn{3}{c}{DBP-WD} & \multicolumn{3}{c}{DBP-YG} \\ 
		\cline{2-4} \cline{5-7} & Hits@1 & MR & MRR & Hits@1 & MR & MRR \\
		\hline 
			Rel. view   & 67.59 & 301  & 0.726 & 12.71 & 3,504 & 0.174 \\
			Attr. view  & 49.08 & 986  & 0.551 & 57.78 & 701  & 0.630 \\ 
			MultiKE-ITC & 83.98 & 421  & 0.866 & 58.46 & 732  & 0.636 \\
		\hline
	\end{tabular}}
\caption{Results of unsupervised entity alignment with MultiKE-ITC}
\label{tab:unsuperv}
\end{table}
\paragraph{Comparison with Conventional Methods.}
We compared MultiKE with a famous and open-source conventional entity alignment method LogMap (version 2.4) \cite{LogMap}. In Table~\ref{tab:logmap}, we depict the results measured by the conventional precision, recall and F1-score. For embedding-based entity alignment, recall and F1-score are equal to precision, because the embedding-based methods always return a list of candidates for every input entity. LogMap is very competitive. In fact, it outperformed other comparative methods in Table~\ref{tab:compare}. MultiKE achieved better results than LogMap, which again demonstrated its effectiveness and practicability. 

\begin{table}[!t]
\centering
\resizebox{\columnwidth}{!}{\scriptsize
	\begin{tabular}{l|cccccc}
		\hline & \multicolumn{3}{c}{DBP-WD} & \multicolumn{3}{c}{DBP-YG} \\ 
		\cline{2-4} \cline{5-7} & Prec. & Recall & F1-score & Prec. & Recall & F1-score \\
		\hline LogMap & 86.61 & 80.68 & 83.54 & 82.71 & 83.73 & 83.22 \\
			  MultiKE & \textbf{91.45} & \textbf{91.45} & \textbf{91.45} & \textbf{88.03} & \textbf{88.03} & \textbf{88.03} \\
		\hline
	\end{tabular}}
\caption{Comparison with LogMap}
\label{tab:logmap}
\end{table}


\section{Conclusion and Future Work}
\label{sect:concl}

In this paper, we proposed a multi-view KG embedding framework for entity alignment, which learns entity embeddings from three representative views of KGs. We introduced two cross-KG training methods for alignment inference. We also designed three kinds of strategies to combine view-specific embeddings together. Our experiments on two real-world datasets demonstrated the effectiveness of our framework. In future work, we plan to investigate more feasible views (e.g., entity types) and study cross-lingual entity alignment.

\section*{Acknowledgments}
This work is supported by the National Natural Science Foundation of China (Nos. 61872172, 61772264) and the Collaborative Innovation Center of Novel Software Technology and Industrialization.

\bibliographystyle{named}
\bibliography{ijcai19}

\end{document}